\documentclass[conference]{IEEEtran}
\IEEEoverridecommandlockouts
\usepackage{cite}
\usepackage{amsmath,amssymb,amsfonts}
\usepackage{graphicx}
\usepackage{textcomp}
\usepackage{xcolor}
\usepackage{times}  
\usepackage{helvet}  
\usepackage{courier}  
\usepackage{url}  
\usepackage{graphicx,color}  
\usepackage{stackengine}

\usepackage{booktabs}
\usepackage{etoolbox}
\usepackage{algorithm}
\usepackage[noend]{algpseudocode}
\usepackage{lipsum,multicol,times, epsfig, graphicx, amsmath, amssymb,  booktabs, url, multirow, comment, subfigure, mwe}
\usepackage{textcomp}
\usepackage{caption}
\usepackage{verbatim}

\usepackage[font=small,skip=1pt]{caption}

\graphicspath{ {./images/} }
\usepackage[T1]{fontenc}

\def\BibTeX{{\rm B\kern-.05em{\sc i\kern-.025em b}\kern-.08em
    T\kern-.1667em\lower.7ex\hbox{E}\kern-.125emX}}
\begin{document}

\title{MAPEL: Multi-Agent Pursuer-Evader Learning using Situation Report}

\author{\IEEEauthorblockN{Sagar Verma}
\IEEEauthorblockA{CVN, CentraleSup\'elec, Universit\'e Paris-Saclay\\
sagar.verma@centralesupelec.fr}
\and
\IEEEauthorblockN{Richa Verma}
\IEEEauthorblockA{TCS Innovation Lab\\
richa15054@iiitd.ac.in}
\and
\IEEEauthorblockN{P.B. Sujit}
\IEEEauthorblockA{IIIT Delhi\\
sujit@iiitd.ac.in}
}

\maketitle

\begin{abstract}
In this paper,  we consider a territory guarding game involving pursuers, evaders and a target in an environment that contains obstacles.  The goal of the evaders is to capture the target, while that of the pursuers is to capture the evaders before they reach the target. All the agents have limited sensing range and can only detect each other when they are in their observation space. We focus on the challenge of effective cooperation between agents of a team. Finding exact solutions for such multi-agent systems is difficult because of the inherent complexity.  We present Multi-Agent Pursuer-Evader Learning (MAPEL), a class of algorithms that use spatio-temporal graph representation to learn structured cooperation. The key concept is that the learning takes place in a decentralized manner and agents use situation report updates to learn about the whole environment from each others' partial observations. We use Recurrent Neural Networks (RNNs) to parameterize the spatio-temporal graph. An agent in MAPEL only updates all the other agents if an opponent or the target is inside its observation space by using situation report. We present two methods for cooperation via situation report update: a) Peer-to-Peer Situation Report (P2PSR) and b) Ring Situation Report (RSR). We present a detailed analysis of how these two cooperation methods perform when the number of agents in the game are increased. We provide empirical results to show how agents cooperate under these two methods.
\end{abstract}


\section{Introduction}
\begin{figure}[t]
    \centering
        \stackunder[5pt]{\includegraphics[scale=0.20]{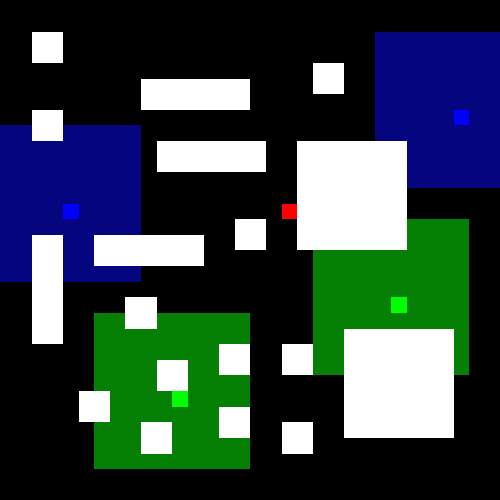}}{(a) Game snapshot}
        \hspace{2mm}
         \stackunder[5pt]{\includegraphics[scale=0.28]{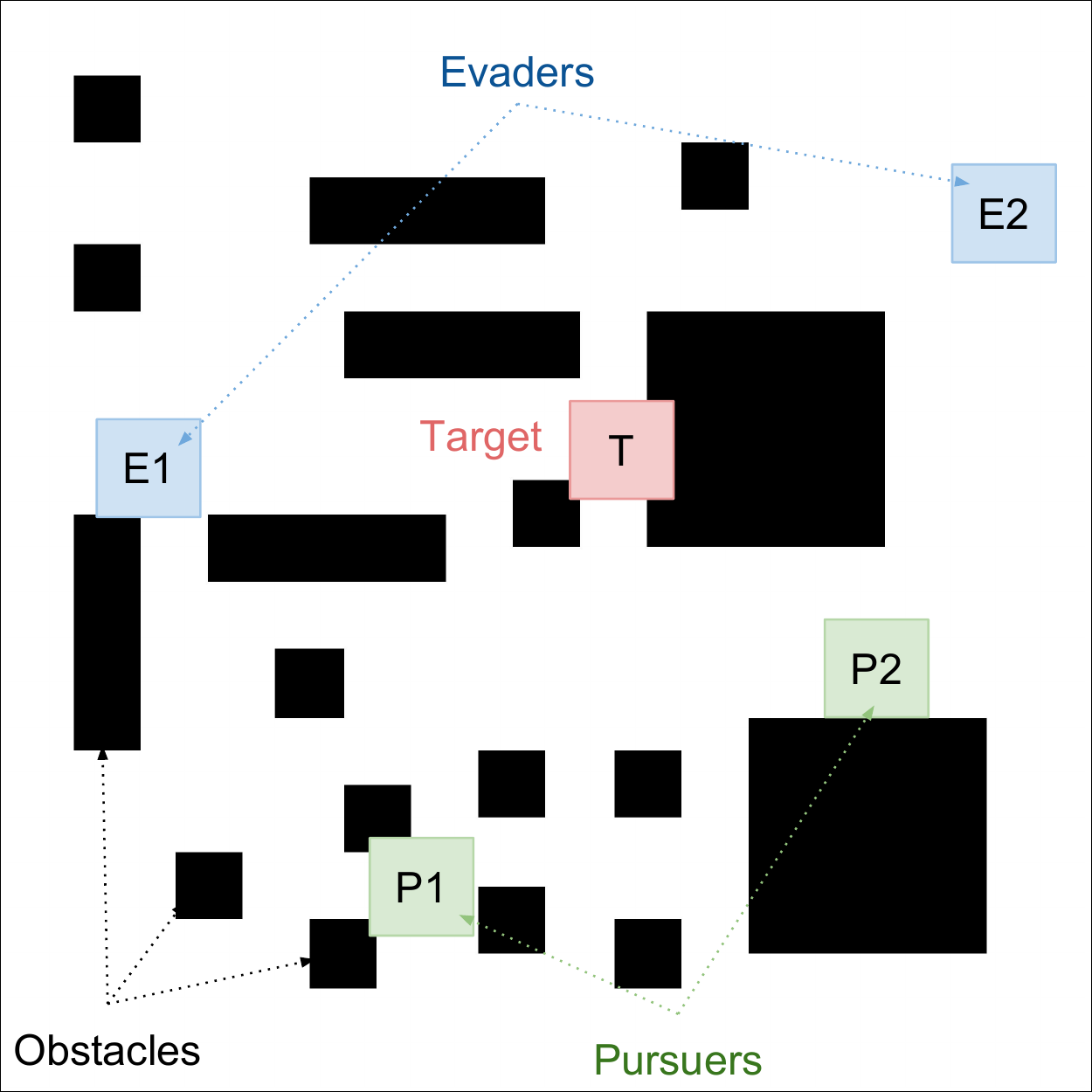}}{(b) Scene description} \\
        \vspace{1mm}
        \stackunder[5pt]{\includegraphics[width=0.41\textwidth, height=35mm]{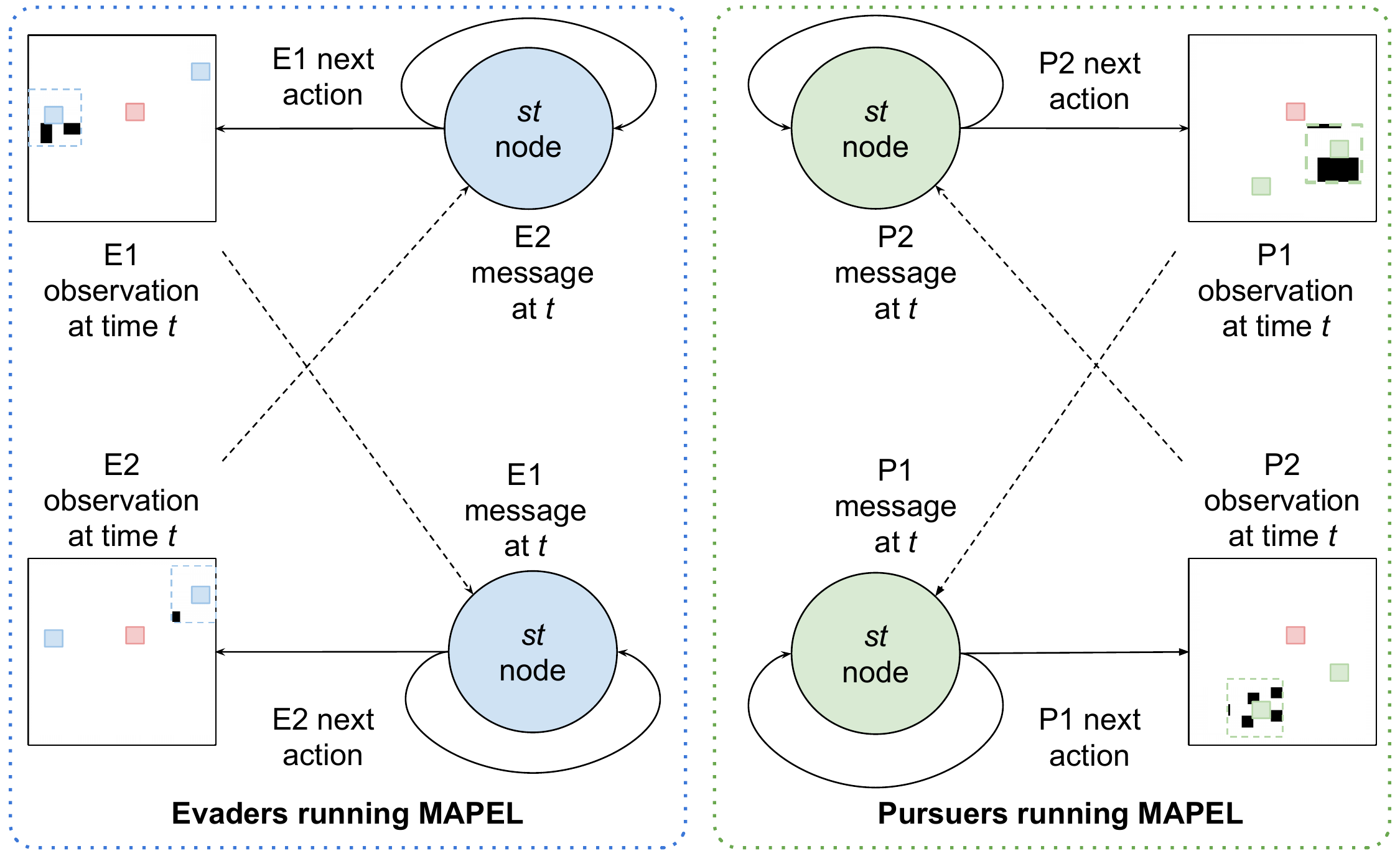}}{(c) Proposed method} \\
        \vspace{1mm}
     \caption{Top-left image shows the rendering of our multi-agent pursuer-evader game. Top-right image shows the labelled description of the game environment. Bottom image shows the proposed cooperation method.}
     \label{fig:teaser}
\end{figure}

Multi-agent systems have received much attention in the past decade \cite{kuo2015, pmlr-v48-he16, Littman1994, MirowskiPVSBBDG16}. In such systems, agents share a common environment, where they act independently or in cooperation with each other to achieve a combined goal. We focus on the problem where multiple agents achieve a single task in cooperation with each other. In such problems, the agents must have the ability to handle unknown and uncertain scenarios and take the success of the whole team into account.

A multi-agent pursuit-evasion game complexity depends on many variables like the type of environment, the observation of the agents, their actions, cooperation strategy, and the reward structure. Being complex and dynamic, pursuit-evasion problems are challenging to solve \cite{antoniades2003}. Such complexities have been addressed by stochastic modeling of agents motion in \cite{Hespanha01greedycontrol, GUIBAS1999}. There has been growing interest in modeling the game, in which the evader is intelligent and has certain sensing capabilities \cite{vidal2002}. This paper focuses on the problem of partial observation of agents where structured message passing is used for cooperation.

We present a zero-sum game based on pursuit and evasion between two teams of equal number agents. Since either the pursuer or the evader wins the game, therefore, the game can be represented as a zero-sum game. The multi-agent pursuit-evasion problem is shown in Figure \ref{fig:teaser}. We assume partial observability of the environment to ensure that the solution is usable in many real-world applications that closely correspond to the task in hand. However, learning becomes more difficult under partial observation along with complex interactions between agents and the environment. Each agent perceives the environment locally and even though the effect of other agents' actions on the environment is visible but the agents, themselves, are not. Reinforcement learning (RL) has been used to solve multi-agent pursuit-evasion in \cite{Parker2002,Yong2009,bilgin2015,kuo2015}. Recent works like \cite{Zhang2018,lowe2017multi,hong2018aamas,Leibo2017} use deep reinforcement learning for different multi-agent problems where centralized policy learning is employed. All these works deal with full observation and are not suitable for our problem. We propose MAPEL, a class of deep reinforcement learning based methods that uses spatio-temporal graphs for structured cooperation between agents under partial observation.

To solve our multi-agent pursuit-evasion game, we present MAPEL which uses spatio-temporal graphs to structure the cooperation between agents in a team. We propose using abstract messages called situation reports which are shared among agents for cooperation. We present two different methods for situation report update which are based on dense and sparse communication. MAPEL can handle pursuers and evaders which move at the same speed throughout the game, which means that neither pursuers nor evaders have an advantage over each other. We show that MAPEL cooperation methods lead to a high degree of cooperation between agents. We also show how the two cooperation methods perform when the number of agents in the team is increased.

The remainder of this paper is organized as follows. Section 2 mentions the existing works related to this paper. Section 3 describes the problem and its formulation as a multi-agent reinforcement learning (MARL) problem. Section 4 presents MAPEL and other proposed baselines. Section 5 explains the experimental setup followed by the results and their explanation in section 6. Section 7 concludes this paper.

\section{Related Work}

Reinforcement learning has been successfully used to play games like Atari \cite{mnih2013playing} and Go \cite{silver2017mastering}. In \cite{liu2009pursuit}, the authors suggest an approach based on hierarchical RL for the same, while enabling the players to learn through tasks with less complexity. Multi-agent reinforcement learning (MARL) consists of a set of learning agents that share a common environment \cite{busoniu2008comprehensive}. Learning in such a framework is fundamentally difficult because of the interaction arising between the agents and the environment and amongst themselves. Conventional decentralized learning techniques like Q learning for each agent \cite{tan1993multi} assume the other agents to be a part of the environment. Such methods don't work in multi-agent settings because the theoretical convergence guarantee no longer holds and makes the learning unstable due to the fact that changes in the policy of any agent will affect the policies of the other agents, as well \cite{matignon2012independent}.

Joint action learning or centralized policy learning is one way to do multi-agent reinforcement learning. \cite{hong2018aamas} present a deep policy inference Q-network that targets multi-agent systems composed of controllable agents. A centralized policy for the controllable agent is learned from its raw observations. \cite{OmidshafieiPAHV17} presents joint and independent policy learning methods. In an independent policy learning method, the joint learned policy is transferred to individual agents in an iterative manner. \cite{Gupta2017} discusses why centralized policy learning fails in case of multi-agent setting and presents methods to learn policy for heterogeneous agents as well as homogeneous agents. Sunehag et al. \cite{Sunehag2018} discuss the problem of "lazy agents" which is when some agents remain inactive when a centralized policy learning is used. They present a value-decomposition network which enables better reward sharing between agents to solve the problem of inactive agents.

Decentralized learning requires effective cooperation between different agents. \cite{Foerster2018aamas} suggest learning with opponent-learning awareness method in which each agent anticipates other agent's policy. This method only works for complete observation. \cite{Palmer2018aamas} discusses the problem of experience replay in multi-agent deep reinforcement learning (MA-DRL). They state that transitions stored in experience replay memory (ERM) can become outdated because agents update their policies in parallel. They apply leniency to MA-DRL by mapping agents state-action pairs to decaying temperature values that control the amount of leniency applied towards negative policy updates that are sampled from the ERM. They also state that this help in better cooperation among agents. \cite{lowe2017multi} use actor-critic to learn policies for complex cooperation. \cite{FoersterFANW17} uses centralized critic to estimate the Q-value, decentralized actors are used to optimize agents' policies, and counterfactual baselines are used to solve multi-agent credit assignment problem. \cite{Yong2001} presents co-evolution methods to learn better coordination between agents. \cite{Pinheiro2018aamas} present a method to solve cooperation between agents that can act selfishly.

Some multi-agent problems can be explicitly described as graphs. \cite{Shao2017} presents cooperative reinforcement learning for multiple agents in StarCraft game. \cite{Hu2010} uses graph representation with reinforcement learning for coordination and cooperation in multi-agent patrol task. Efficient state representation based on the distance between agents and different game entities is used to reduce the observation state complexity. \cite{Marzag2017aamas} presents a flag coordination game where graph structure is explicitly present and is utilized to model multi-agent coordination. Some problems where there are no explicit graph structures present, game states can be decomposed into some weak time-varying structures. Such structures can be learned using factor graph representation and graphical learning methods. \cite{Bryant2018} discusses use cases where cooperation is explicitly required. A genetic algorithm variation is used to solve the adaptive team of agents (ATA) problem. Their method can adapt an agent to a new role based on the overall structure of the environment. \cite{Guestrin2002} presents joint policy learning method for coordinated reinforcement learning through structured communication between agents. \cite{Zhang2014} presents a way to decompose a global Q-function into local Q-function based on the task decomposition between agents expressed using factor graphs. \cite{Zhang2013} divide agents into cliques based on specific tasks. \cite{amato2015} uses factor graphs to learn implicit structures present in multi-agent settings. Factor graphs reduce the action and observation space and learning agents' policy becomes easier.

 In the literature, RL has been used earlier for the classic pursuer-evader game\cite{isaacs1999differential}. In \cite{liu2010novel}, a learning technique for multi-player pursuit-evasion games is presented for discrete state and action spaces. The proposed algorithm is only applicable for multi-player pursuit-evasion games with superior pursuers (in terms of speed). The article \cite{wang2015research} is another work suggesting a technique using learning in differential multi-player pursuit-evasion games that have superior evaders. In \cite{alexopoulos2015iros} hierarchical decomposition is used to solve games having two pursuers and one evader.

\section{Problem Definition}
\label{sec:probdef}

We model the multi-agent pursuer-evader problem as a grid world of dimension $M \times N$ in which obstacles are placed randomly (uniform distribution $\mathcal{N}(0,\sigma)$). In this grid, there are $P$ pursuers, $E$ evaders, and a single target $T$. At any time $t$, a pursuer $p\in P$ has the global knowledge about all pursuer locations and the current target location in the environment. An evader $e\in E$ is assumed to know the locations of all other evaders and the target. We assume each agent can sense a rectangular region of length $l$ and width $w$. However, the agents cannot sense on the other side of the obstacle. That is, a pursuer can detect an evader if they are in line-of-sight and within the sensed region. 
The speed of all the pursuers and the evaders is given by $v$ and remains constant throughout the game. The target, $T$ remains stationary throughout the game.

A game starts with randomly sized obstacles placed on the grid at random locations as shown in  Figure \ref{fig:teaser} (for a 2-pursuers vs 2-evaders game). The target is spawned at a random location near the middle of the grid ($M/2,N/2$) and it is of length $t_s$. The pursuers and the evaders are randomly spawned on the opposite sides of the grid. 
The pursuers and the evaders can move to any of the adjacent cells of the grid only if the cell is either empty or occupied by any of the agents. An agent reaches the target when its location is same as the target's location. Also, a pursuer captures an evader only if their locations are the same. Once an evader is captured by a pursuer, it cannot move anywhere else but the pursuer can move to an adjacent cell after catching the evader.

There are three conditions for a game to complete.
\begin{enumerate}
    \item An evader reaches the target, in which case the evaders win the game.
    \item A pursuer reaches the target before an evader, in which case the pursuers win the game.
    \item All the evaders are captured by the pursuers, in which case the pursuers win the game.
\end{enumerate}

Based on the three different winning criteria we have the following reward structure:
\begin{enumerate}
    \item When the evaders win by capturing the target, a reward of $w^e=0.5$ is awarded to them and a penalty of $w^p=-0.5$ is given to the pursuers.
    \item When the pursuers win by reaching the target before the evaders, a reward of $w^p=0.5$ is awarded to, and a penalty of $w^e=-0.5$ is given to the evaders.
    \item When the pursuers win by capturing all the evaders, a reward of $w^p=1$ is awarded to the pursuers and a penalty of $w^e=-1$ is given to the pursuers.
\end{enumerate}

Rewards are equally divided among all the agents of a team. This makes it sure that agents in a team do not compete with each other.

\section{Methods}

In this section we first present a naive method in which an agent greedily moves towards the target, followed by the second method which is a multi-agent formulation of deep Q-learning and then we introduce the proposed method MAPEL with two different cooperation strategies.

\subsection{Naive Method}
A naive agent tries to move towards the target, $T$. Each agent has a partial view of the environment and knows the location of the other agents of its team. It also knows the location of the target. A naive agent moves towards the target in a straight line. If the next location on the line of sight towards the target is obstructed, it randomly chooses an adjacent location that is closest to the line. If a pursuer observes an evader in its observation space, it computes the shortest path to the evader and chooses its next location along that path. Similarly, if a pursuer/evader observes the target in its observation space, it computes the shortest path to the target and chooses its next location along that path.  Also, if a pursuer observes the target and an evader or multiple evaders in its field of view, it computes the shortest paths to all of them and chooses its next location along the path that has the smallest length.

\subsection{Multi-agent Q-learning}
\begin{figure*}[t]
    \centering
        \includegraphics[scale=0.47]{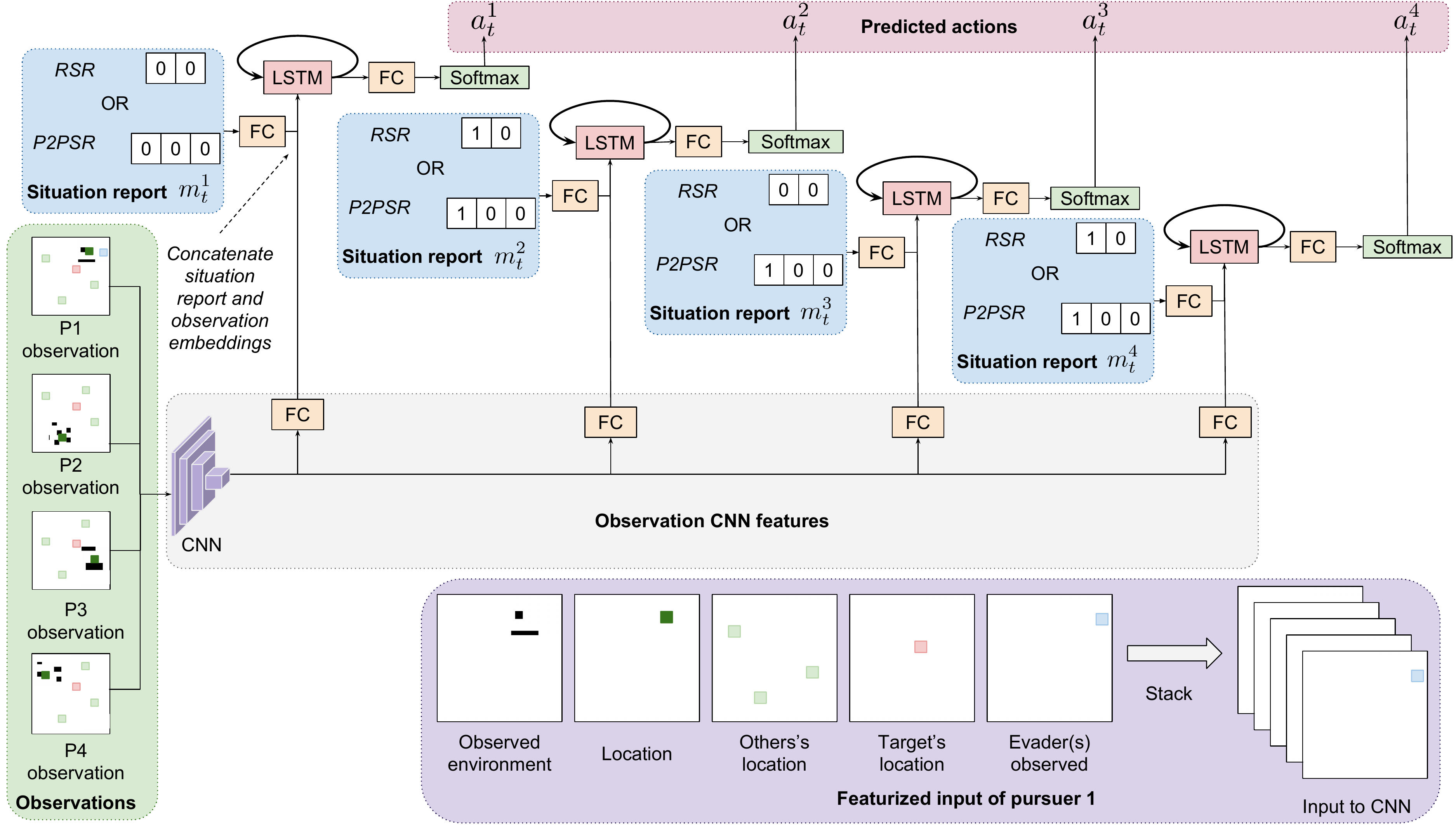} \\
        \vspace{1mm}
     \caption{Generalized architecture of MAPEL.}
     \label{fig:architecture}
\end{figure*}

An $N$-agent stochastic game $\mathcal{E}$ is formalized by the tuple $\mathcal{E} = (\mathcal{S},\mathcal{A}^1,\dots,\mathcal{A}^N,\mathcal{R}^1,\dots,\mathcal{R}^N,\mathcal{T},\gamma)$, where $\mathcal{S}$ denotes the state space, and $\mathcal{A}^i$ is the action space of agent $j\epsilon\{1,\dots,N\}$. The reward function for agent $j$ is defined as $\mathcal{R}^i:\mathcal{S}\times\mathcal{A}^1\times\dots \times\mathcal{A}^N \rightarrow \mathbb{R}$, determining the immediate reward. The transition probability is given by $\mathcal{T}:\mathcal{S}\times\mathcal{A}^1\times\dots\mathcal{A}^N \rightarrow Pr(\mathcal{S})$. $Pr(\mathcal{S})$ is the collection of probability distributions over the state space $\mathcal{S}$.  The goal of agents is to find a policy $\pi$ which maximizes the expected return $G_t$, which is the discounted sum of rewards given by $G_t = \sum_{i=1}^{N}\sum_{\tau=t}^{T}\gamma^{\tau-t}\mathcal{R}^{i}_{\tau}$, where $T$ is the time-step when an episode ends, $t$ denotes the current time-step, $\gamma \epsilon [0,1)$ represents the reward discount factor, and $\mathcal{R}^{i}_{\tau}$ is the reward received at time-step $\tau$ by agent $\mathcal{A}^i$.

The agents choose actions according to their policies. For agent $i$, the corresponding policy is defined as $\pi^{i}:\mathcal{S}\rightarrow Pr(\mathcal{A}^i)$, where $Pr(\mathcal{A}^i)$ is the collection of probability distributions over agent $i$'s action space $\mathcal{A}^i$. The joint policy of all the agents is given by $\pi : \pi^1\times,\dots\times\pi^N$. The joint actions of all the agents is given by $a : \mathcal{A}^1\times,\dots,\times\mathcal{A}^N$. The value function of agent $i$ given state $s$ under the joint policy $\pi$ is written as the expected cumulative discounted future reward:

\begin{equation}
    v^i_{\pi}(s) = v^i(s;\pi)=\sum_{t=0}^{\infty}\gamma^t\mathbb{E}_{\pi,p} \big[r^i_t|s_0=s,\pi\big]
    \label{eq:vfunc}
\end{equation}

The $Q$-function can then be defined within the framework of $N$-agent game based on the Bellman equation given the value function in equation \eqref{eq:vfunc} such that the $Q$-function $Q^i_{\pi}:\mathcal{S}\times\mathcal{A}^1\times,\dots,\mathcal{A}^N\rightarrow \mathbb{R}$ of agent $i$ under the joint policy $\pi$ can be formulated as

\begin{equation}
    Q^i_{\pi} = R^i(s,a) + \gamma\mathbb{E}_{s'-p}\big[v^i_{\pi}(s')\big],
    \label{eq:qfunc}
\end{equation}
where $s'$ is the state at the next time step. The value function $v^i_{\pi}$ can be expressed in terms of the $Q$-function in equation \eqref{eq:qfunc} as

\begin{equation}
    v^i_{\pi} = \mathbb{E}_{a~\pi}\big[Q^j_{\pi}(s,a)].
    \label{eq:vfunc2}
\end{equation}

The $\mathcal{Q}$-function for $N$-agent game in equation \eqref{eq:qfunc} extends the formulation for a single-agent game by considering the joint action taken by all agents $a$, and by taking the expectation over the joint action in equation \eqref{eq:vfunc2}.

\subsection{Multi-Agent Pursuer-Evader Learning (MAPEL)}

In the Q-learning method presented in the previous section, the joint policy $\pi$ is dependent only on the current observation of all the agents combined. It is impossible for an agent to know anything about the observation of the other agents in that setting. Also, the size of the combined observation and action spaces increases exponentially with the number of agents. For a large number of agents, this could be problematic.

\begin{figure}[t]
    \centering
        \stackunder[6pt]{\includegraphics[scale=0.32]{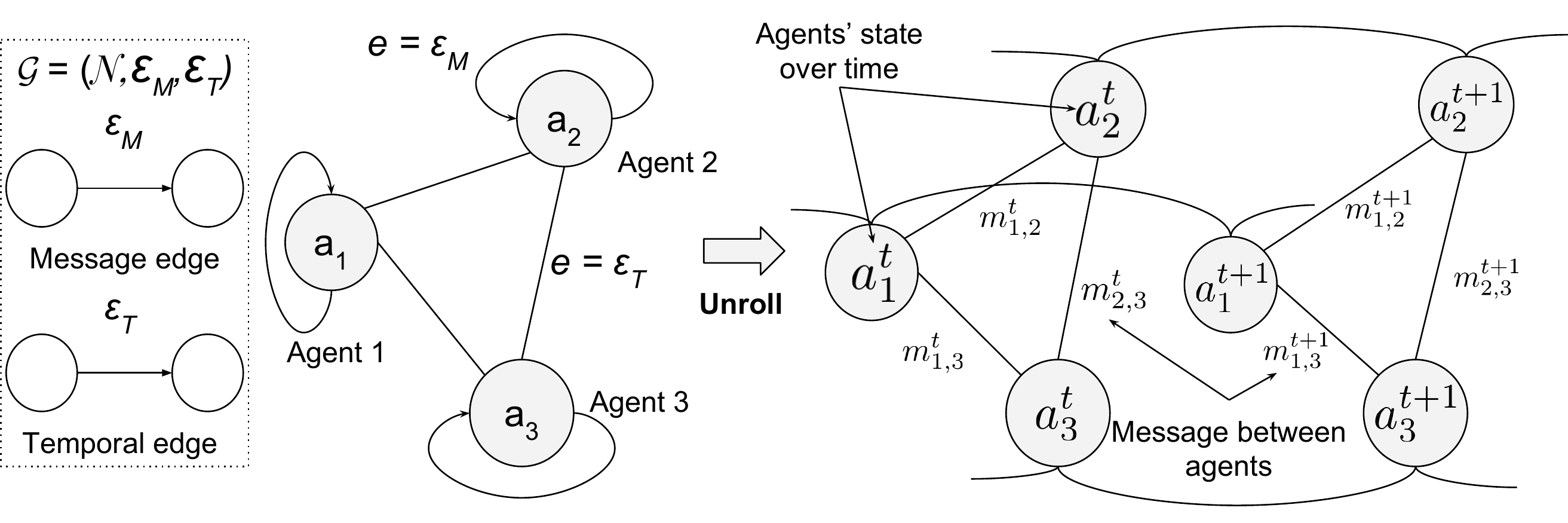}}{(a) Spatio-temporal representation}\\
        \vspace{1mm}
        \stackunder[6pt]{\includegraphics[scale=0.35]{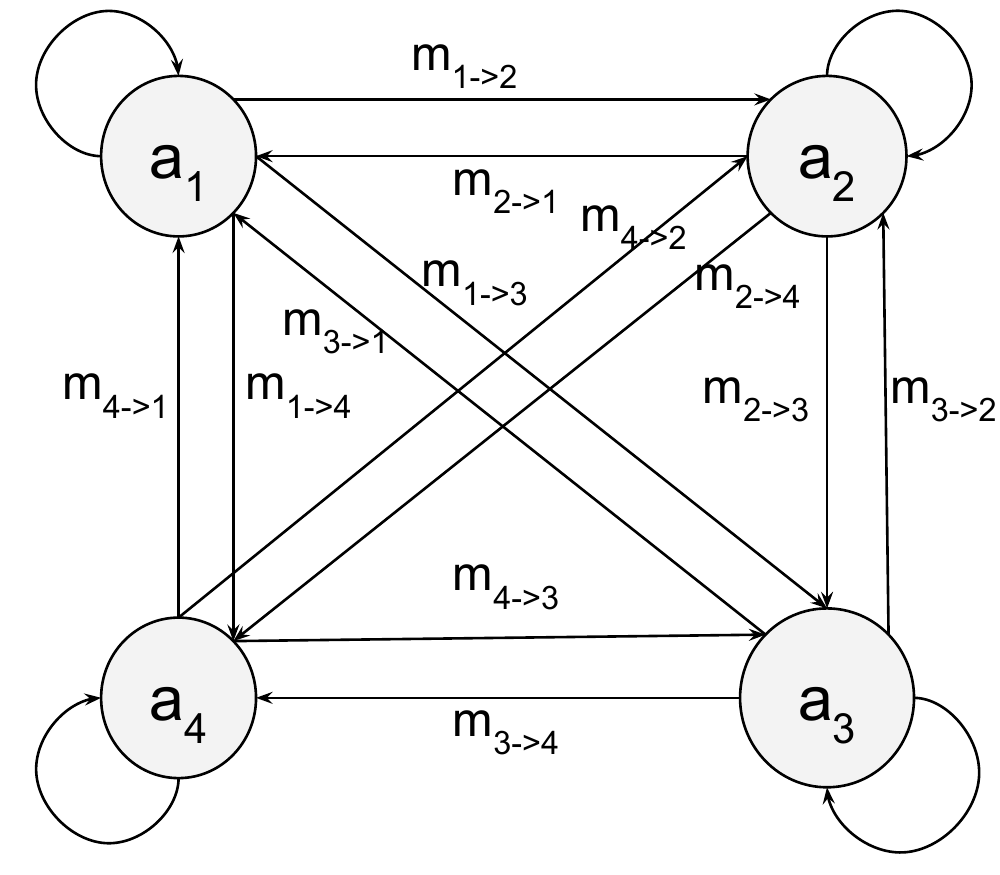}}{(b) P2PSR} \hspace{5mm}
        \stackunder[6pt]{\includegraphics[scale=0.35]{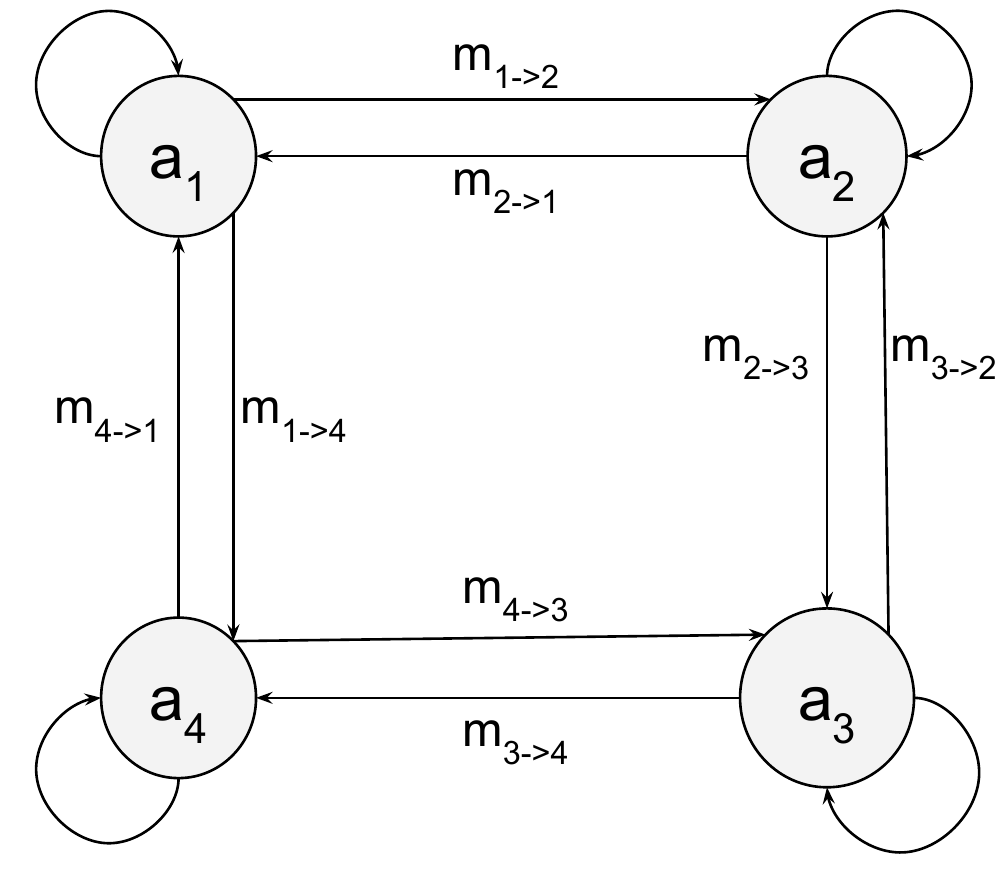}}{(c) RSR}\\
        \vspace{1mm}
     \caption{Spatio-temporal representation used for cooperation between agents. Top image shows cooperation between three agents using spatio-temporal graph unrolled in time. Bottom-left image shows peer-to-peer situation report method between four agents and bottom-right image shows ring situation report method between four agents.}
     \label{fig:st-graph}
\end{figure}

In this section, we present a spatio-temporal (st) architecture called MAPEL which allows agents to learn their individual policies by sharing their observations with each other by cooperating via situation reports. We represent a team of agents as an st-graph $\mathcal{G} = (\mathcal{N},\mathcal{E}_M,\mathcal{E}_T)$, where $\mathcal{N}$ denotes the total number of agents, $\mathcal{E}_M$ is the total number of edges between the agents i.e. the edges used to pass situation reports, and $\mathcal{E}_T$ is the number of edges connecting agents at time $T$. Figure \ref{fig:st-graph} shows an example st-graph capturing agent-agent interactions during a game. In the unrolled st-graph, two agents at a given time step $t$ are connected with an undirected \textit{spatio-temporal} edge $e=(a_i,a_j) \epsilon \mathcal{E}_M$, and two nodes at adjacent time steps are connected with an undirected \textit{temporal} edge \textit{iff} $(a_i,a_j)\epsilon\mathcal{E}_T$.

We parameterize the nodes $\mathcal{N}$ and edges $\mathcal{E}_T$ using RNNs in our st-graph. The edges $\mathcal{E}_M$ are used by nodes to pass situation reports to each other. The situation reports are used by agents to compute their actions at time $t$. The network architecture of MAPEL is illustrated in figure \ref{fig:architecture}, each agent is represented by an RNN, the agents compute their observational features using a CNN and pass their own observations to other agents via situation reports. Each agent uses the situation reports received by other agents along with its current observation to compute its next action. RNN maintains history information about an agent. The situation report coupled with RNN is used to handle partial observability. Situation report provides an abstract and clear representation of the observation. This helps in reducing the hidden state representation noise which arises due to other agents changing their strategies. For example, if a pursuer observes the target or an evader, it can inform other pursuers about its observation via situation report. This could help the other pursuers in changing their decision to not go in the direction of this particular pursuer and search in other areas for the target or other evaders.

In real-world applications, we can have hundreds of agents and interaction among all of them may not be possible due to some physical constraints or simply because of high computational complexity. In most of the cases, it is not necessary to have dense communication between all the agents. Sparse communication structures can be used to learn effective cooperation. We present two situation report update methods that use the structures present in the game.

\paragraph*{\textbf{Peer-to-Peer Situation Report (P2PSR)}}

In Peer-to-Peer Situation Report method, all the agents can share situation report with each other. This is the case of dense communication. This means that in our st-graph representation for $\mathcal{N}$ nodes, we have $\mathcal{E}_M=\mathcal{N}(\mathcal{N}-1)/2$ edges. Figure \ref{fig:st-graph} shows the st-graph representation of P2PSR. This type of cooperation is required when an agent wants to know what other agents are observing so that it does not explore their regions. The objective of the agents then becomes to minimize the search time and exploration area to complete the task.

\paragraph*{\textbf{Ring Situation Report (RSR)}}

In Ring Situation Report method, agents are randomly chosen to form a ring. Each agent can only pass messages to its adjacent agents. The st-graph representation for this type of cooperation is given by Figure \ref{fig:st-graph}. For $\mathcal{N}$ nodes, we have $\mathcal{E}_M= \mathcal{N}$ edges for $\mathcal{N}>2$. This type of cooperation can be used to cordon off an area and search inside it. This does not require all the agents to know about the other agents' observations. An agent only needs to know what its adjacent agents are observing. This decreases the number of messages required to cooperate.

\section{Experimental Setup}

\begin{figure}[t]
    \centering
        \stackunder[6pt]{\includegraphics[scale=0.26]{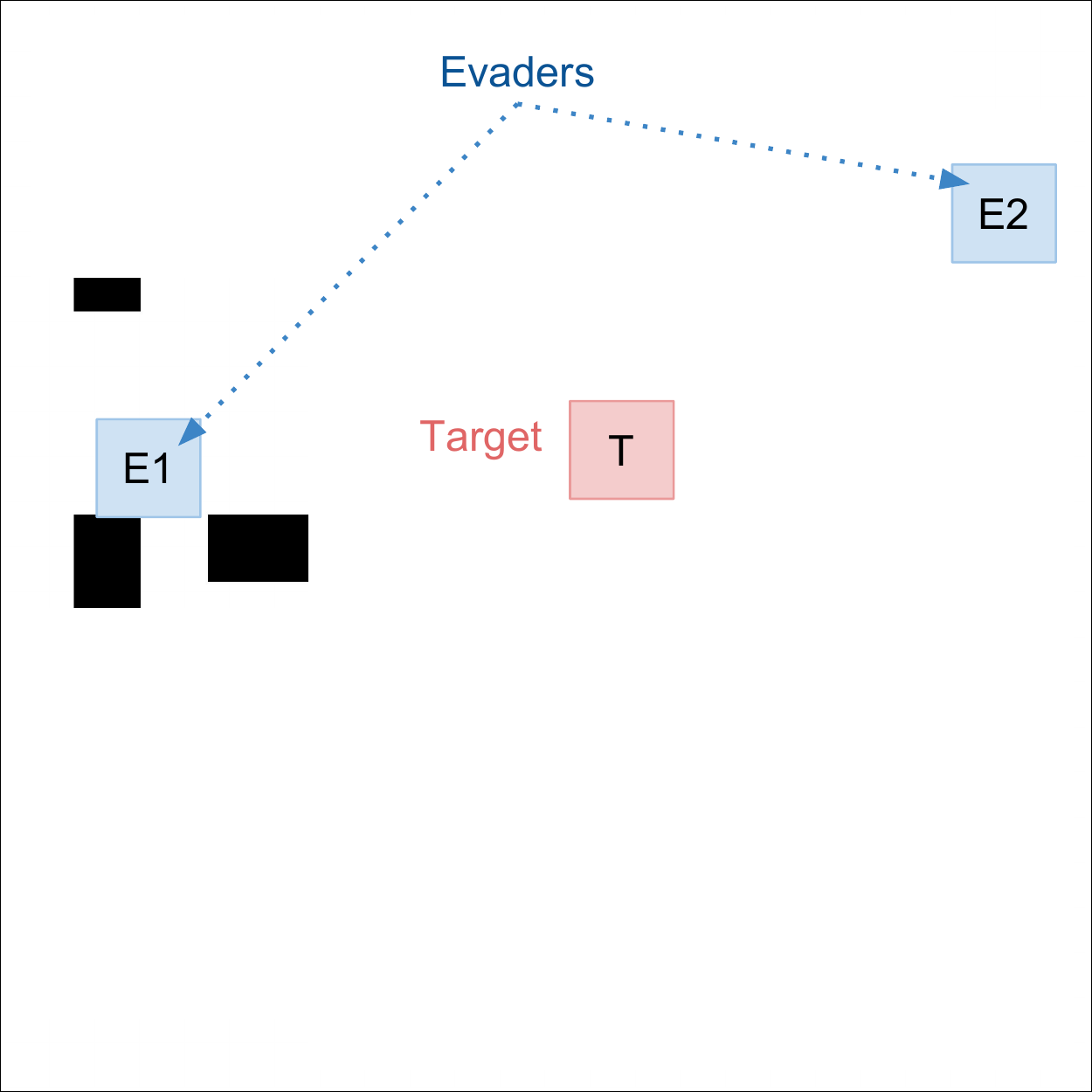}}{(a) Evader1 observation}
        \hspace{2mm}
        \stackunder[6pt]{\includegraphics[scale=0.26]{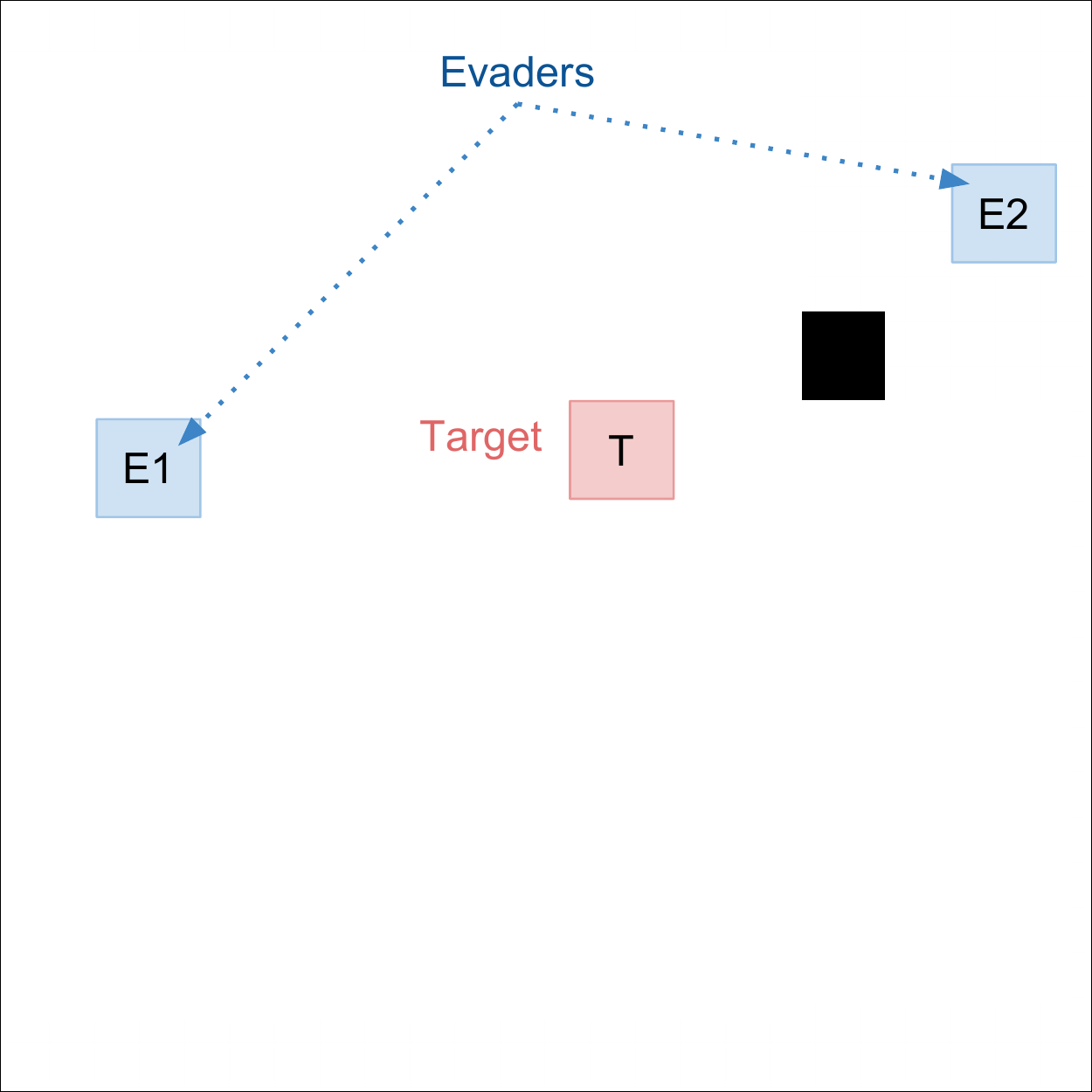}}{(b) Evader2 observation} \\
        \vspace{1mm}
        \stackunder[6pt]{\includegraphics[scale=0.26]{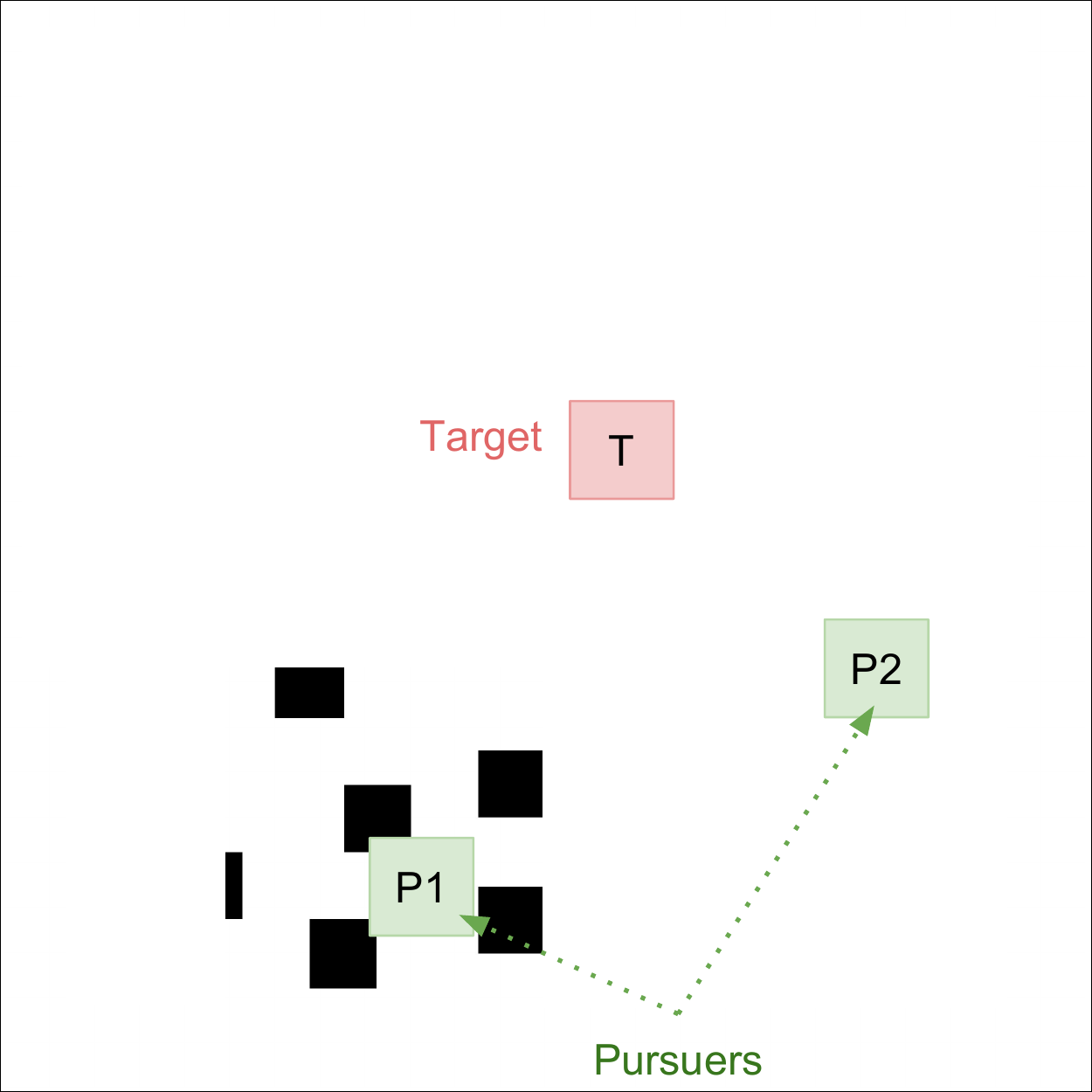}}{(c) Pursuer1 observation} \hspace{2mm}
        \stackunder[6pt]{\includegraphics[scale=0.26]{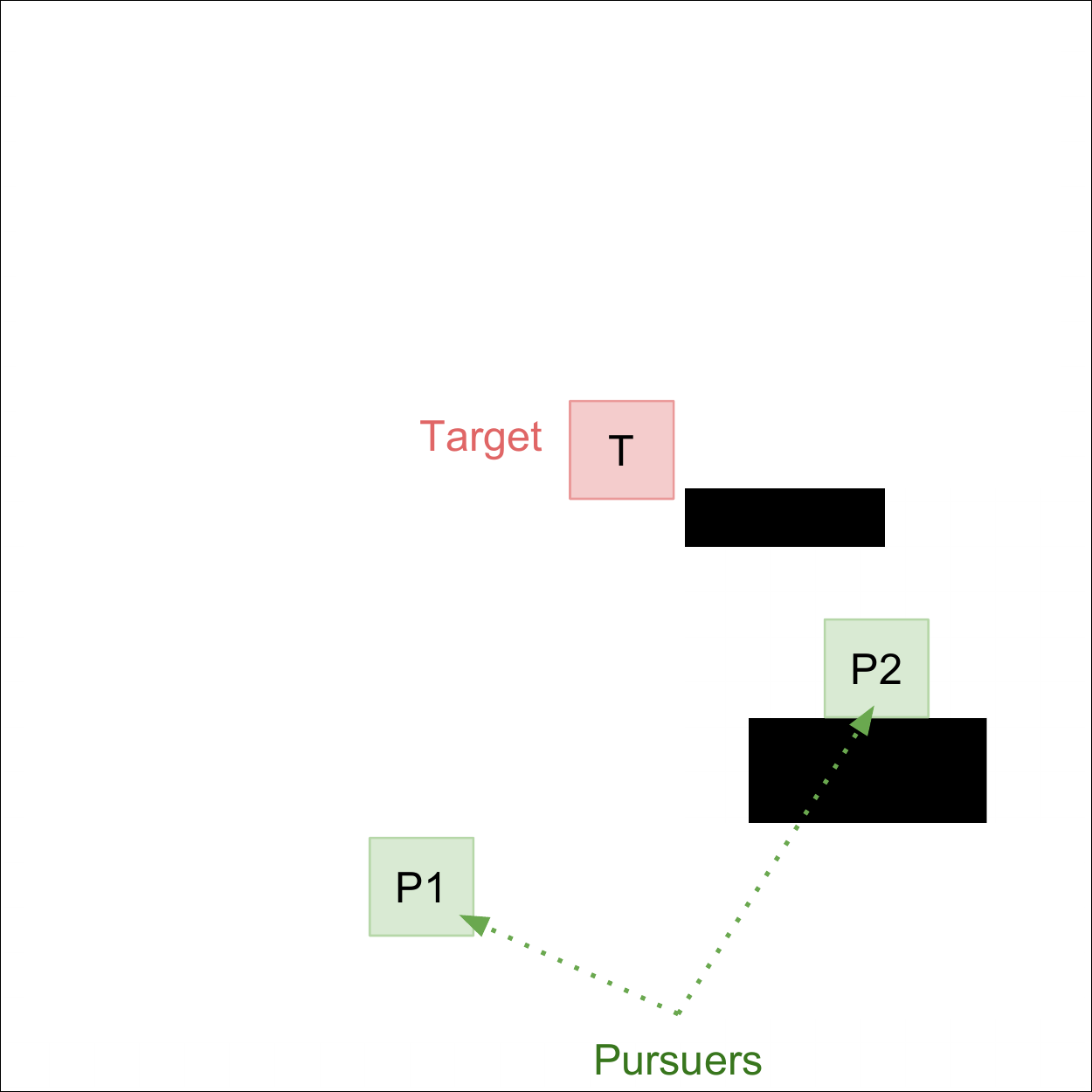}}{(d) Pursuer2 observation}\\
        \vspace{1mm}
     \caption{Partial observation of individual agents in 2-pursuers vs 2-evaders game. Each agent can see everything in its observation space and knows about the target and other team members' locations. An agent cannot see the observation of its team members.}
     \label{fig:partial_obs}
\end{figure}

We perform our experiments on the multi-agent pursuer-evader environment presented in section \ref{sec:probdef}. We begin with explaining the environment representation, agent observation featurization, and representation of messages under different situation report methods. All the experiments have been conducted on a workstation with 1.2 GHz CPU, 256 GB RAM, NVIDIA V100 GPU and running Ubuntu 18.04. We use PyTorch \cite{paszke2017automatic} for network implementation.

\subsection{Environment}

The environment is a grid world composed of multiple grids of size $32\times32$. Figure \ref{fig:partial_obs} shows partial observation of agent in a 2 vs 2 game. The white regions are the empty regions where evaders and pursuers can move. An agent considers all grid cells outside its observation space to be empty. Evaders are blue, pursuers are green, and the target is red. Figure \ref{fig:partial_obs} (a) shows the observation space of evader $1$, it can see all the grid cells in its observation space, it knows the locations of other evaders and the target. Similarly figure \ref{fig:partial_obs} (b) shows the observation space of evader $2$. Figure \ref{fig:partial_obs} (c) shows the observation space of pursuer $1$, it can see all the grid cells in its observation space, it knows the locations of other pursuers and the target. Similarly, the observation space of pursuer $2$ is represented in figure \ref{fig:partial_obs} (d).

\subsection{Agent Observation}

For $Q$-learning, we need to represent an agent's observation as meaningful features. In our experiments, we found that raw RGB frames provide good observational for 2 vs 2 games but fail to generalize for more number of agents. We represent each type of entity in our environment as separate channels. We have five channels in our feature space, each of size $32\times32$. In the bottom left portion of figure \ref{fig:architecture}, we show the featurization of observation of one of the pursuers in a 4 vs 4 game. The first channel shows the observation space of the agent, the second channel shows the position of the agent itself, the third channel shows the position of other agents, the fourth one shows the location of the target, and the fifth channel shows the location of the opponent(s) observed. This feature representation accurately incorporates all the information observed by an agent.
\begin{figure*}[t]
    \centering
        \stackunder[6pt]{\includegraphics[scale=0.4]{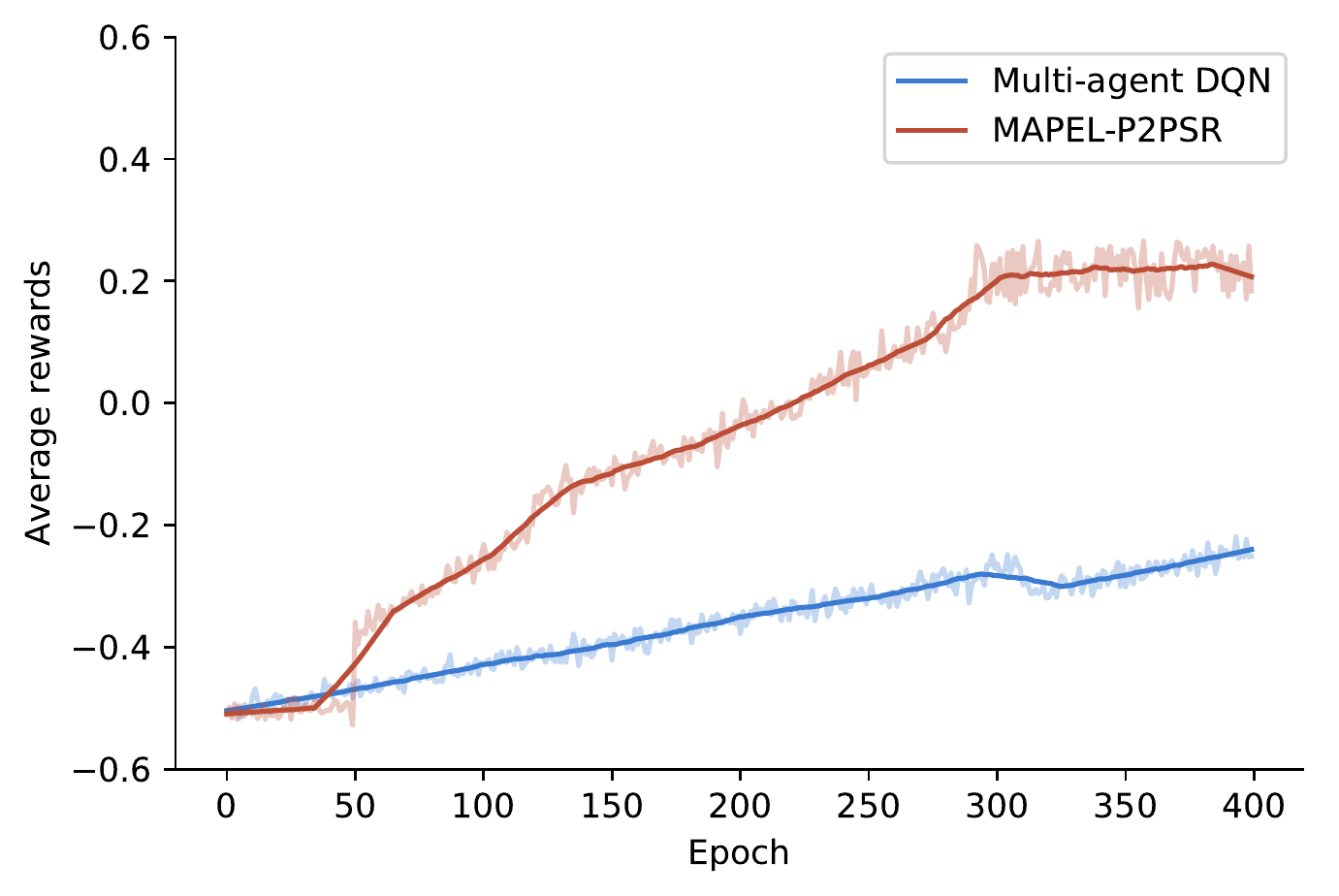}}{(a) 2 vs. 2, against naive pursuers}
         \hspace{2mm}
         \stackunder[6pt]{\includegraphics[scale=0.4]{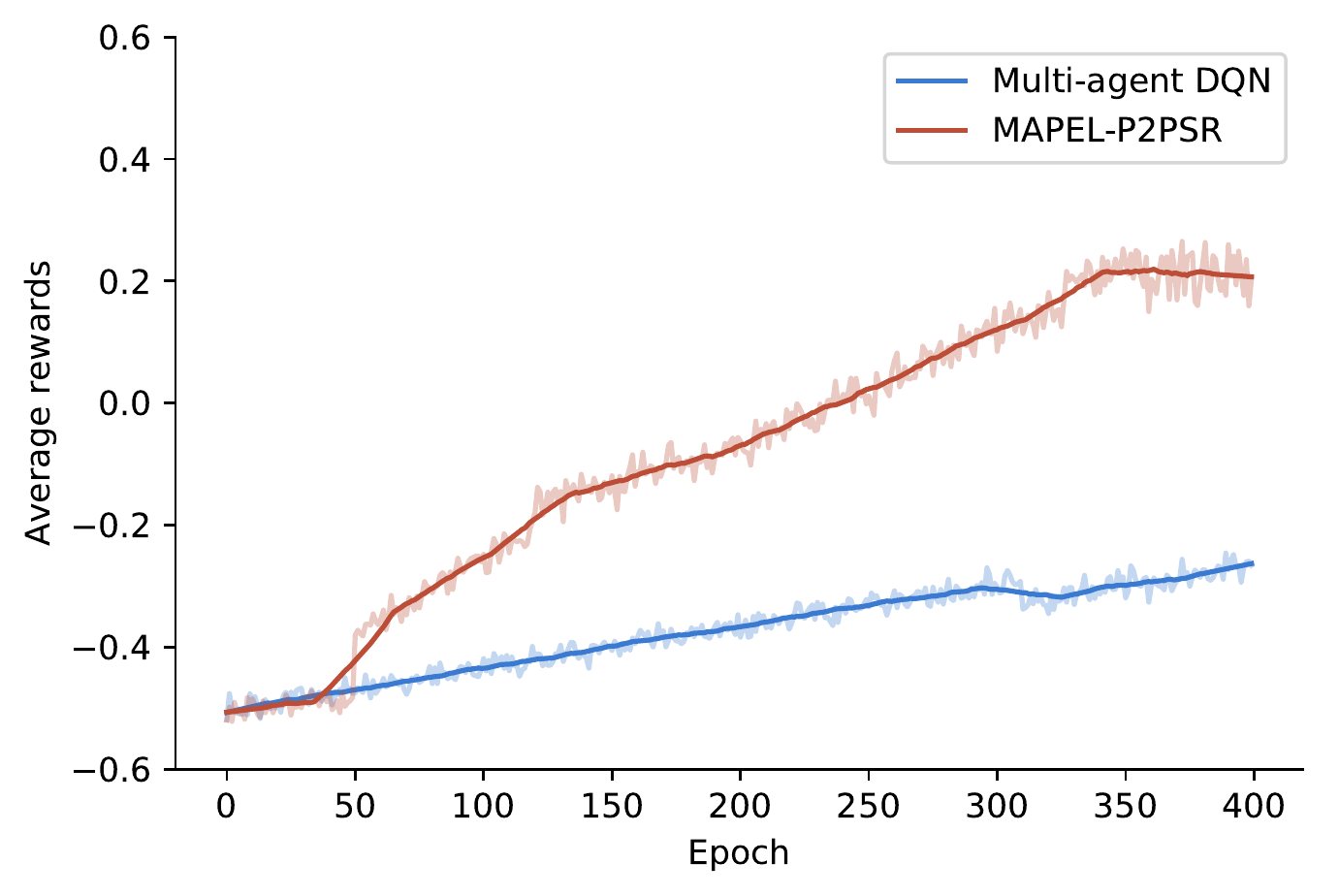}}{(b) 3 vs. 3, against naive pursuers}
         \hspace{2mm}
        \stackunder[6pt]{\includegraphics[scale=0.4]{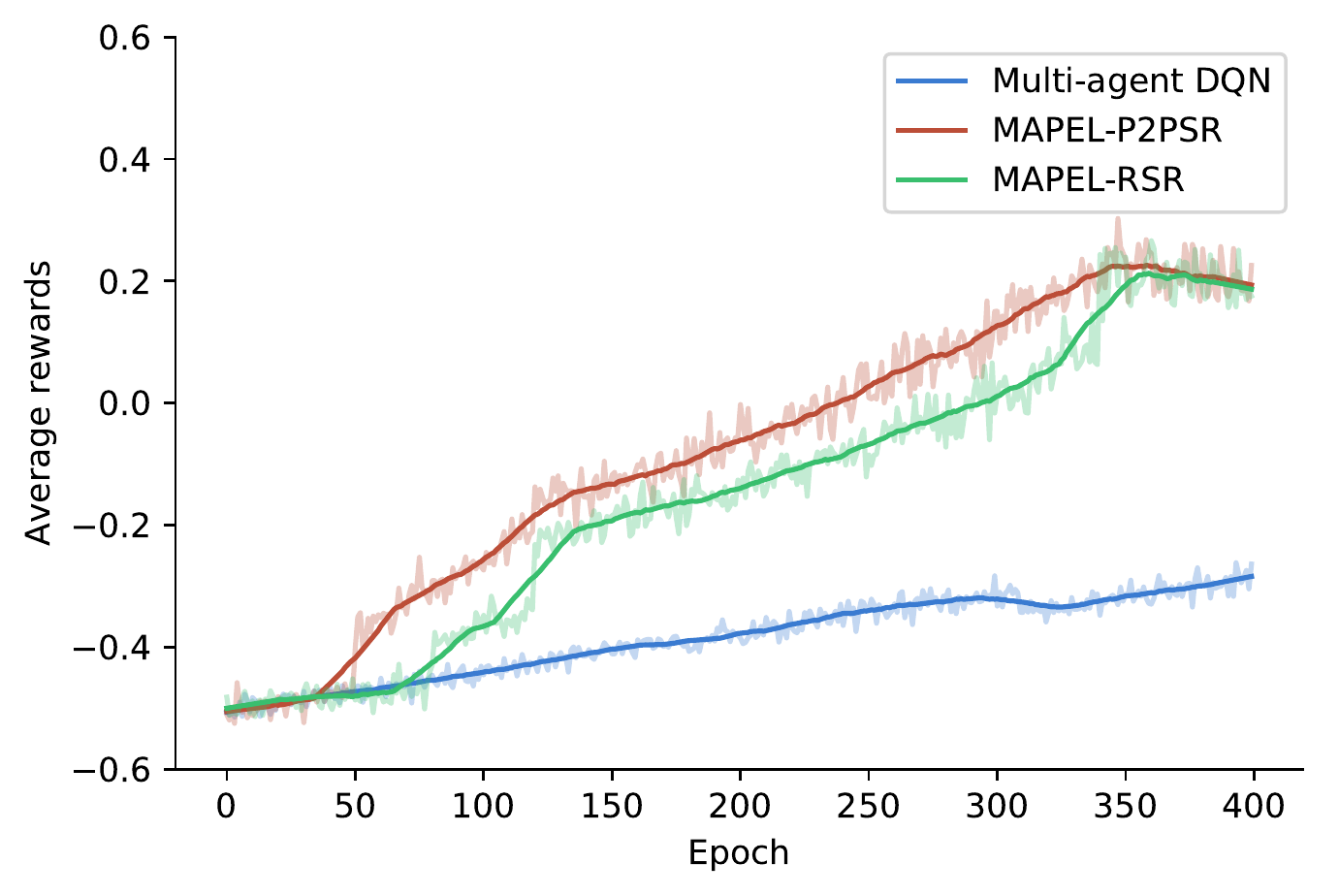}}{(c) 5 vs. 5, against naive pursuers} \hspace{2mm}\\
        \vspace{1mm}
        \stackunder[6pt]{\includegraphics[scale=0.4]{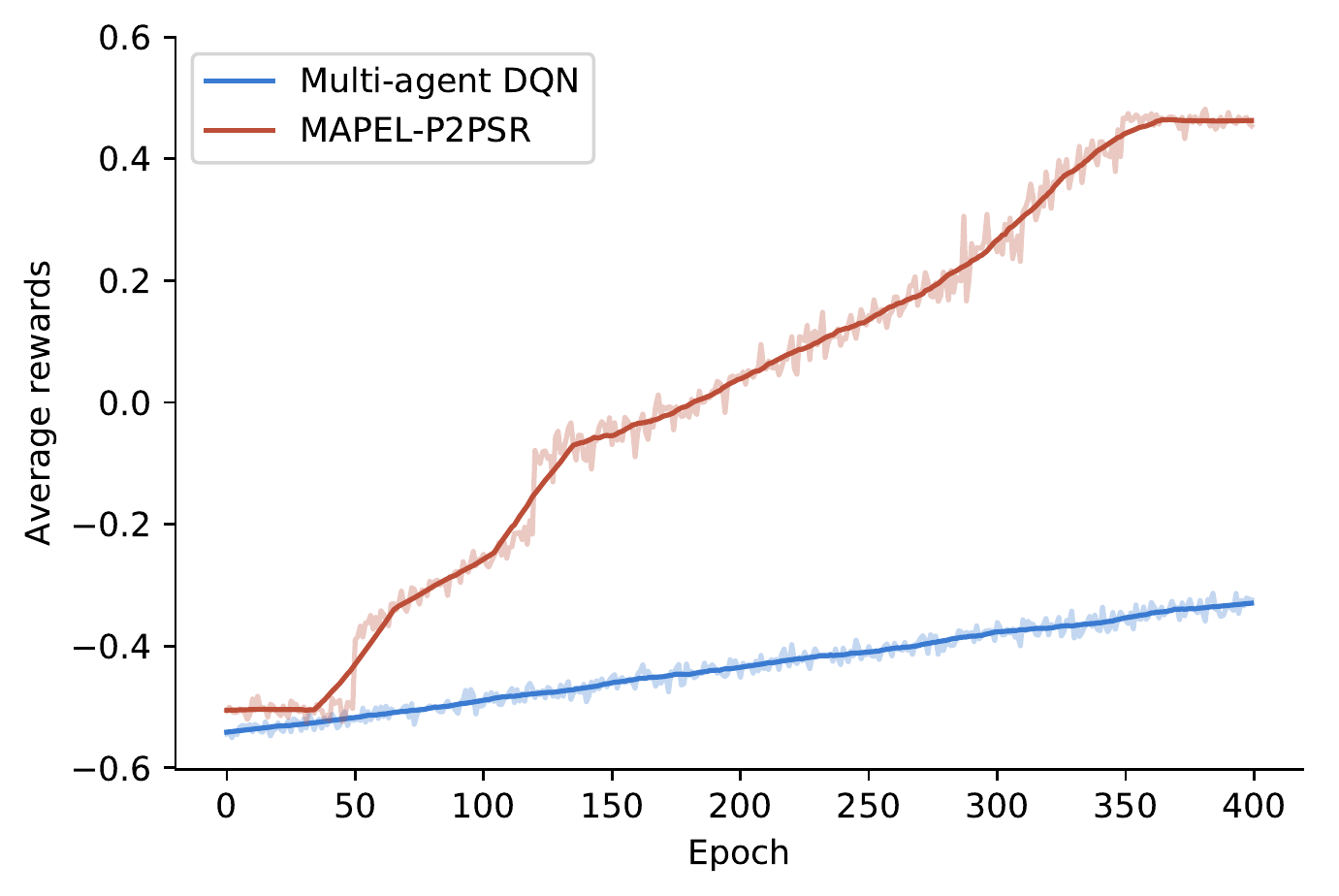}}{(d) 2 vs. 2, against naive evaders}
         \hspace{2mm}
         \stackunder[6pt]{\includegraphics[scale=0.4]{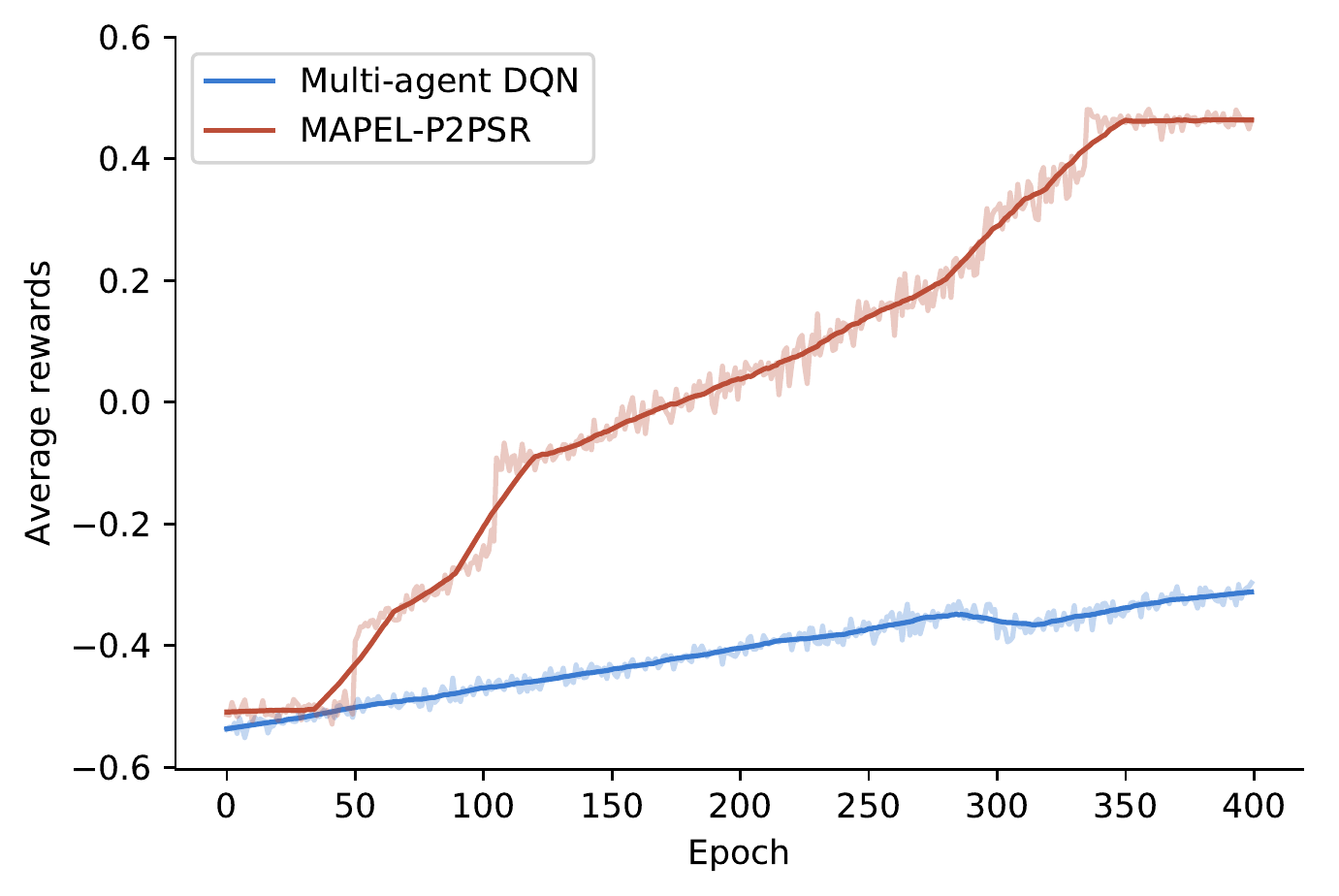}}{(e) 3 vs. 3, against naive evaders}
         \hspace{2mm}
        \stackunder[6pt]{\includegraphics[scale=0.4]{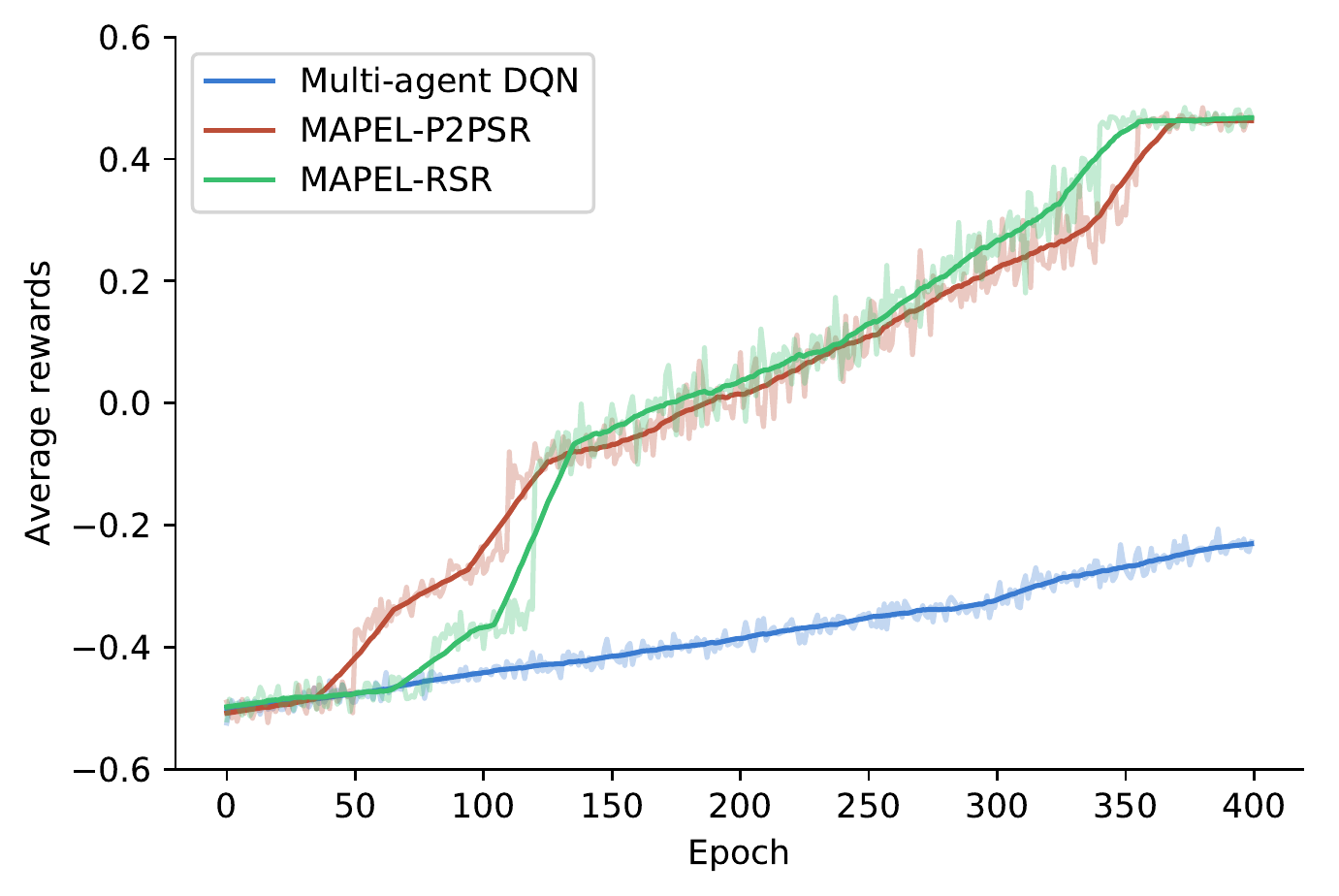}}{(f) 5 vs. 5, against naive evaders} \hspace{2mm}\\
        \vspace{1mm}
     \caption{Learning curve comparison of different methods in different scenarios.}
     \label{fig:converge}
\end{figure*}

\subsection{Message Representation}

In P2PSR, an agent $a_i$ receives a situation report $m^{t}_i$ in form of a vector of size $\mathcal{N}-1$ at time $t$. The $\mathcal{N}-1$ elements of the vector represent the messages from other agents, if the value of an element in the vector is $1$, then it means the corresponding agent has seen the target or opponent(s) in its observation space. An element with value $0$ means that the observation space is empty. In RSR, an agent $a_i$ receives a situation report message $m^{t}_i$ in form of a vector with size $2$ from two of its adjacent agents at time $t$.

\subsection{Training}

We train multi-agent DQN for learning evaders against naive pursuers and multi-agent DQN for learning pursuers against naive evaders. We train both MAPEL cooperation methods against naive agents. All our models are trained for 400 epochs, 500 episodes per epoch. We use Adam \cite{KingmaB2014ICLR} optimizer to train all our models. Learning rate is varied over epochs, it starts with 0.001 and decays at every 200 epochs by one-tenth. To ensure exploration, $\epsilon$-greedy starts at 1.0 and ends at 0.1. A discount factor of 0.99 is used. While training multi-agent DQN models, a history length of 5 observations is used. We use a batch size of 64 in all our experiments. We vary the number of agents per team from 2 to 5 for all our models. Following are the model variations that we train,

\begin{enumerate}
\item MA-DQN pursuers against naive evaders.
\item MA-DQN evaders against naive pursuers.
\item MAPEL-P2PSR pursuers against naive evaders.
\item MAPEL-P2PSR evaders against naive pursuers.
\item MAPEL-RSR pursuers against naive evaders.
\item MAPEL-RSR evaders against naive pursuers.
\end{enumerate}

\subsection{Evaluation}

\begin{table*}[]
    \centering
    \begin{tabular}{c  c c  c c  c c c c}
        \toprule
         & \multicolumn{2}{c}{Naive} & \multicolumn{2}{c}{MA-DQN} & \multicolumn{2}{c}{MAPEL-P2PSR} & \multicolumn{2}{c}{MAPEL-RSR} \\
         Scenario & Average & Complete & Average & Complete & Average & Complete & Average & Complete\\
         & reward & wins & reward & wins & reward & wins & reward & wins\\
         \midrule
         2 vs. 2 &   0.159 & 9.77\% &   -0.274 & 3.13\% & 0.431 & 14.62\% & NA & NA \\
         3 vs. 3 &   0.161 & 10.23\% &  -0.235 & 3.17\% & 0.396 & 15.79\% & NA & NA \\
         4 vs. 4 &   0.162 & 10.07\% &  -0.217 & 2.92\% & 0.479 & 16.71\% & 0.456 & 15.92\% \\
         5 vs. 5 &   0.165 & 10.13\% &  -0.213 & 2.72\% & 0.483 & 16.23\% & 0.468 & 15.92\% \\
         \bottomrule
    \end{tabular}
    \caption{Evaluation result of different methods for pursuers against naive evaders in different scenarios.}
    \label{tab:allpursuer}
\end{table*}

\begin{table*}[]
    \centering
    \begin{tabular}{c  c   c   c  c}
        \toprule
         Scenario & Naive & MA-DQN & MAPEL-P2PSR & MAPEL-RSR \\
         \midrule
         2 vs. 2 &   0.134  &  -0.279 &  0.419  & NA  \\
         3 vs. 3 &   0.153  &  -0.247 &  0.429  & NA  \\
         4 vs. 4 &   0.157  &  -0.225 &  0.423  & 0.416  \\
         5 vs. 5 &   0.161  &  -0.217 &  0.417  & 0.419  \\
         \bottomrule
    \end{tabular}
    \caption{Evaluation result of different methods for evaders against naive pursuers in different scenarios.}
    \label{tab:allevader}
\end{table*}

We evaluate our MA-DQN, MAPEL-P2PSR, and MAPEL-RSR evaders against naive pursuers and vice-versa. 100,000 episodes are used for all the evaluations. Average reward is reported for both evaders and pursuers. In the case of pursuers running different methods, we also report the total number of times pursuers were able to capture all the evaders. We call these results as "complete wins".

\section{Results}

Figure \ref{fig:converge} compares different models' learning curve under the different number of agents for both evaders and pursuers. For the number of agents $\mathcal{N}={1,2,3}$, both MAPEL cooperation methods, i.e., P2PSR and RSPRP have the same message length. In such cases, there is no fundamental difference between these methods. Therefore, we only train both methods when team sizes are more than 3. Figure \ref{fig:converge} (a) shows the learning curve for evaders with MA-DQN and MAPEL-P2PSR when the number of agents is 2 for both evaders and pursuers. Similarly figure \ref{fig:converge} (b) is for a 3 vs. 3 scenario for evaders against naive pursuers. Figure \ref{fig:converge} (d) and (e) are for pursuers with MA-DQN and MAPEL-P2PSR against naive evaders when the numbers of agents are 2 and 3 respectively. Figure \ref{fig:converge} (c) and (f) show all three methods for evaders and pursuers against their naive opponents when the number of agents is 5.

In all the scenarios, pursuers are able to score better rewards than evaders. We believe that the pursuers are able to learn about the strategy where capturing all the evaders maximizes their rewards. From figure \ref{fig:converge} (c) and (f), it is evident that MAPEL-P2PSR for pursuers learns about capturing all evaders quickly as compared to MAPEL-RSR. After 350 epochs both the methods converge to same average rewards which shows that MAPEL-RSR has similar learning capabilities as MAPEL-P2PSR. We believe this is due to the fact that in the case of MAPEL-P2PSR, all pursuers know about all other pursuers' observations explicitly which helps them in knowing about "capture all evaders" strategy early. In the case of MAPEL-RSR, more epochs are required to learn about this strategy.

Table \ref{tab:allpursuer} compares the average rewards and complete wins of different methods for pursuers against naive evaders in four scenarios, i.e., 2 vs. 2, 3 vs. 3, 4 vs. 4, and 5 vs. 5. It can be seen that the naive method performs better than MA-DQN in all the scenarios. On rendering a few episodes, we find that MA-DQN pursuers are not able to find the shortest paths as compared to the naive method. For some of the successful episodes, we find that pursuers are able to beat the opponents when some of the team members are closer to the target as compared to the evaders. In 4 vs 4 and 5 vs 5 scenarios, we can see that MAPEL-P2PSR is ahead of MAPEL-RSR by 0.023 and 0.025 units of average reward respectively. This is in line with our earlier hypothesis that MAPEL-P2PSR is better at learning about "capture all evaders" strategy because of dense communication. This is evident from the "complete wins" in column 7 and 9. The difference in the average reward is still less when compared to difference in "complete wins" between the two methods.

Table \ref{tab:allevader} compares the average rewards and complete wins of different methods for evaders against naive pursuers in four scenarios, i.e., 2 vs. 2, 3 vs. 3, 4 vs. 4, and 5 vs. 5. Similar to the case of evaders, the naive method performs better than MA-DQN in all the scenarios. We also observe that the rewards from MAPEL-P2PSR and MAPEL-RSR for evader are smaller than the pursuers. The reason for this is that pursuers can learn about "capture all evaders" strategy to get more reward whereas pursuers don't have any such strategy to maximize their rewards further. The reason MAPEL methods perform better than naive and MA-DQN methods is that evaders can avoid the regions where pursuers have been observed by some members of the team.

\section{Conclusions and Future Work}
In this paper, we presented a variation of multi-agent pursuit-evasion game with partial observability. We also present MAPEL for multi-agent cooperative reinforcement learning to solve the game. We compare proposed MAPEL with two benchmarks; the naive method which is a greedy solution and a multi-agent DQN formulation. We perform experiments with varying number of agents to show the generalizability of the MAPEL cooperation methods. We empirically show that MAPEL cooperation methods are better at learning cooperation strategy by reporting the results of "capture all evaders" in the case of pursuers.

In the future, our goal would be to test the transfer-ability of MAPEL methods to games with more number of agents. We would also like to experiment under different game conditions like opponents with different speeds, non-equal team sizes, moving target, etc. We would also like to find effective ways of analyzing and comparing proposed cooperation methods.

\vspace{2mm}

\bibliographystyle{IEEEbib}  
\bibliography{ms}  

\end{document}